\documentclass[final,5p,times,twocolumn,authoryear]{elsarticle}

\usepackage{amssymb}
\usepackage{lipsum}
\usepackage{amsmath}
\usepackage{array}
\usepackage{tabularx}
\usepackage{supertabular}
\usepackage{pdflscape}
\usepackage{rotating}
\usepackage{threeparttable}
\usepackage{booktabs}
\usepackage{longtable}
\usepackage{url} 
\usepackage{tikz}
\usepackage{hyperref}
\usepackage{array}
\usepackage{float} 
\usepackage{pifont}   
\usepackage{amssymb}  
\usepackage[authoryear]{natbib}  

\usetikzlibrary{mindmap,shadows,trees}
\usetikzlibrary{mindmap,trees} 

\usepackage{lineno}

\begin{document}
\begin{frontmatter}


\title{YOLO-Based Pipeline Monitoring in Challenging Visual Environments}

\author[1]{Pragya Dhungana}
\author[2]{Matteo Fresta}
\author[3]{Niraj Tamrakar}
\author[4]{Hariom Dhungana\corref{cor1}}

\affiliation[1]{organization={Nepal Telecommunications Authority}, city={Kathmandu},  country={Nepal}}
            
\affiliation[2]{organization={Department of Electrical Electronics and Telecommunication Engineering and Naval Architecture (DITEN), University of Genoa},  city={Genova},   country={Italy}}

\affiliation[3]{organization={Department of Bio-systems Engineering, Gyeongsang National University}, city={Jinju},
                       country={Korea}}
\affiliation[4]{organization={Department of Mechanical Engineering and Maritime Studies, Western Norway University of Applied Sciences}, city={Bergen},  country={Nepal}}

\cortext[cor1]{}
\ead{hdhu@hvl.no}

\begin{abstract}
Condition monitoring subsea pipelines in low-visibility underwater environments poses significant challenges due to turbidity, light distortion, and image degradation. Traditional visual-based inspection systems often fail to provide reliable data for mapping, object recognition, or defect detection in such conditions. This study explores the integration of advanced artificial intelligence (AI) techniques to enhance image quality, detect pipeline structures, and support autonomous fault diagnosis. This study conducts a comparative analysis of two most robust versions of YOLOv8 and Yolov11 and their three variants tailored for image segmentation tasks in complex and low-visibility subsea environments. Using pipeline inspection datasets captured beneath the seabed, it evaluates model performance in accurately delineating target structures under challenging visual conditions. The results indicated that YOLOv11 outperformed YOLOv8 in overall performance.

\end{abstract}






\end{frontmatter}

\section{Introduction}
\label{introduction}
Subsea pipelines form a critical part of global energy infrastructure, transporting oil, gas, and other resources across vast oceanic distances \citep{Ho2020InspectionPaper}. Ensuring their integrity requires frequent monitoring and inspection, typically carried out by remotely operated vehicles (ROVs) or autonomous underwater vehicles (AUVs). However, underwater environments present unique challenges for visual inspection and data collection due to low light, particulate matter, and variable turbidity, which result in blurred and distorted images \citep{Hasan2018CorrosionDesign}. These visual limitations significantly impact the performance of conventional computer vision algorithms used for fault detection and diagnostic. Recent advances in artificial intelligence offer new possibilities for overcoming these limitations. Deep learning-based image enhancement techniques, robust object detection and image segmentation models are increasingly being applied to improve underwater fault detection and diagnostic. These techniques  extract meaningful information from degraded images, allowing for obstacle avoidance, defect detection, leak detection  \citep{Yang2025AutomaticAlgorithm}.

Corrosion of subsea pipelines poses a serious risk to their structural integrity and operational safety, having led to numerous failures in recent years. This issue is particularly severe in pipelines that lack intelligent monitoring or detection systems, making early identification of corrosion difficult. The internal environment of these pipelines is highly aggressive due to large temperature and pressure variations between the inlet and outlet points. Additionally, the presence of corrosive substances such as carbon dioxide (CO$_2$), hydrogen sulfide (H$_2$S), oxygen (O$_2$), and chloride ions (Cl$^-$) in the transported medium significantly accelerates internal corrosion~\cite{Shao2025ArtificialInformation}.

Condition monitoring, decision-making, and maintenance planning are deeply interrelated aspects of asset management. By collecting real-time data on the health of equipment and infrastructure, condition monitoring enables the early detection of potential issues, which in turn supports informed decision-making to implement timely interventions that enhance reliability and cost-efficiency \citep{Singh2024DevelopmentEquipment}. Biomimicry-inspired decision-making framework for condition monitoring that enhances fault diagnosis through adaptive learning, multi-sensory integration, data augmentation, and energy-efficient sensing methodology was presented in \citep{Dhungana2025BiomimicryMonitoring}. Development and implementation of a fully automated, unmanned subsea pipeline inspection system using an autonomous underwater vehicle (AUV), with shore-based remote control and onboard self-actuating data workflows enabling real-time mission adjustments was presented in \citep{Rumson2021ThePipelines}. Previous studies have demonstrated strong performance of deep learning techniques in fault identification tasks, highlighting their potential for accurate and automated diagnostics \citep{Dhungana2025DeepMonitoring}. Motivated by these results, this work explores the application of a different deep learning model tailored explicitly for subsea pipeline monitoring.  A comparative analysis of YOLOv8 (a one-stage model) and Mask R-CNN (a two-stage model) for instance segmentation in agricultural automation demonstrated YOLOv8’s superior accuracy and faster inference, underscoring its suitability for real-time automated orchard operations \citep{Sapkota2024ComparingEnvironments}. 

Image segmentation is essential for accurately outlining subsea pipelines, enabling precise localization and structural assessment in challenging underwater environments. It facilitates automated monitoring by isolating the pipeline from background noise such as marine growth, sediment, and lighting artifacts. Building on the widespread application of YOLO models, this study aims to systematically compare and evaluate the performance of different versions and variants of YOLO architectures for pipeline segmentation in challenging visual environments. This study aims to provide insights into the suitability, efficiency, and challenges of each model for subsea pipeline monitoring. Moreover, two image enhancement techniques, Contrast Limited Adaptive Histogram Equalization (CLAHE) \citep{Reza2004RealizationEnhancement} and  Dark Channel Prior Dehazing (DCPD) \citep{Lee2016AAlgorithms} for segmentation are also tested, through a series of experiments on SubPipe datasets, we evaluate the effectiveness of the proposed system and demonstrate how AI enhances data readiness for analytical tasks. Our contributions aim to bridge the gap between raw data acquisition and AI model deployment, paving the way for more robust, intelligent, and autonomous data preprocessing pipelines.

This study provides a comprehensive comparative analysis of leading deep learning models for image segmentation, emphasizing their suitability for pipeline outlining,  corrosion/defect detection, event classification. The findings reveal that model selection should be driven by application-specific requirements rather than a one-size-fits-all approach. Specifically, YOLOv11n demonstrates superior performance in real-time, resource-constrained environments, making it highly suitable for lightweight applications such as mobile or embedded systems. These insights aim to guide practitioners in selecting the most appropriate model based on specific use-case demands.

The remainder of this paper is structured as follows:  Section~\ref{sec:DLinPS} reviews the application of deep learning, particularly YOLO models, in pipeline segmentation. Section~\ref{sec:case} presents the case study, including a description of the dataset, data preparation methods, model configurations, computational setup, and evaluation metrics. Section~\ref{sec:result} details the results and discussion, covering the effects of image enhancement, training behavior, validation and testing performance, and a comparative analysis of model variants. Finally, Section~\ref{sec:conclu} concludes the study and outlines potential directions for future work.

\section{Deep learning in Pipeline  segmentation}    \label{sec:DLinPS}
Deep learning has significantly advanced image segmentation tasks, with various architectures such as U-Net, V-Net, LinkNet, FCNs, and FPNs demonstrating strong performance through innovations in network design, data augmentation, and multi-scale feature representation. U-Net employs a contracting-expanding convolutional neural network (CNN) architecture combined with a robust data augmentation strategy, enabling accurate biomedical image segmentation even with a limited number of annotated images \citep{Ronneberger2015U-Net:Segmentation}. V-Net introduces a 3D fully convolutional neural network that performs end-to-end segmentation of volumetric medical images, utilizing a Dice coefficient-based objective function to handle class imbalance and data augmentation for improved performance with limited training data \citep{Milletari2016V-Net:Segmentation}. LinkNet achieves efficient semantic segmentation by reusing encoder features in the decoder through skip connections and a lightweight decoder design, significantly reducing the number of parameters and computations while maintaining competitive accuracy \citep{Chaurasia2017LinkNet:Segmentation}. Fully Convolutional Networks (FCNs) repurpose classification networks into end-to-end trainable models for pixel-wise semantic segmentation by replacing fully connected layers with convolutional layers, enabling them to process arbitrary input sizes and produce corresponding output segmentations efficiently \citep{Long2015FullySegmentation}. Feature Pyramid Networks (FPN) efficiently construct multi-scale feature maps with rich semantics at all levels by leveraging the inherent hierarchical structure of convolutional networks with a top-down architecture and lateral connections, leading to significant improvements in object detection accuracy, especially for detecting objects at different scales \cite{Lin2017FeatureDetection}. Besides these architectures, several advanced DL architectures have been developed for semantic and instance segmentation, each introducing unique innovations to address challenges such as multi-scale object detection, boundary refinement, and real-time inference. Table~\ref{tab:reworks} summarises recent research efforts focused on subsea pipeline inspection, highlighting the diversity of models and objectives employed to address challenges such as defect detection, leak identification, obstacle avoidance, and image enhancement in underwater environments.

\begin{table*}
  \caption{Recent studies for pipeline inspection tasks in subsea environments, demonstrating their effectiveness under varying conditions.}
  \label{tab:reworks}
  \begin{tabular}{llll}
    \toprule
    References & Year & Model & Motivation \\
    \midrule
    \citep{Dang2025EventClassifiers}  & 2025  &  CNN & Event Classification  \\
    \citep{Zhang2021SubseaVehicle}  & 2021  & Classical image processing  &   Leak detection and Obstacle avoidance  \\
    \citep{Tolie2024DICAM:Enhancement}  & 2024  &  Deep Inception and Channel-wise Attention  & Image enhancement     \\
 \citep{Zhang2025PDS-YOLO:Detection}  & 2025  & PDS-YOLOv8n  &  Pipeline Defect Detection \\
 \citep{Yang2025AutomaticAlgorithm}  & 2025  & OFG-YOLO  & Pipeline Defect Detection  \\
 \citep{Zhang2024EP-YOLO:YOLOv7}  & 2024  & EP-YOLO  &  Pipeline Leak Detection \\
 \citep{Wang2024AdvancedDetection}  & 2024  & YOLO-DS  &  Pipeline Defect Detection \\
  \citep{Cui2025Cross-PIC:Pipelines}  & 2025  & Context-learning point cloud segmentation network  &  Pipeline Detection \\
 
  \bottomrule
\end{tabular}
\end{table*}

\textbf{SegFormer:}
SegFormer is a visual transformer model built in PyTorch and trained from scratch \citep{Xie2021SegFormer:Transformers}. SegFormer eliminates the need for handcrafted, computationally intensive components by combining a hierarchical Transformer encoder with a lightweight All-MLP decoder that effectively captures both local and global context through multiscale features. The encoder, known as Mix Transformer (MiT), generates multi-scale feature maps using overlapped patch merging and efficient self-attention with sequence reduction, capturing both local and global contexts without relying on positional embeddings. These features are then processed by the All-MLP decoder, which unifies, upsamples, and fuses them through simple MLP layers to produce the final segmentation mask. SegFormer’s design ensures a large effective receptive field while maintaining high accuracy and computational efficiency, outperforming traditional CNN-based models and prior Transformer-based methods like SETR, particularly in terms of scalability, generalization, and minimal inference overhead.

\textbf{DeepLab:}
DeepLab achieves state-of-the-art semantic image segmentation by employing atrous convolution to control feature resolution and enlarge the receptive field, Atrous Spatial Pyramid Pooling (ASPP) to handle multi-scale objects, and fully connected Conditional Random Fields (CRFs) to refine object boundaries \citep{Chen2018DeepLab:CRFs}. DeepLabV3 is a widely used CNN for image segmentation that employs atrous convolutions to expand the receptive field, enabling more effective detection of objects at multiple scales \citep{Chen2017RethinkingSegmentation}. The model was implemented using the official PyTorch version with weights pre-trained on a subset of the COCO dataset. DeepLabv3+ enhances the DeepLabv3 architecture by incorporating a decoder module and employing atrous separable convolutions within both the encoder's Atrous Spatial Pyramid Pooling and the decoder to improve boundary detail and efficiency in semantic image segmentation \citep{Chen2018Encoder-DecoderSegmentation}.

\textbf{Mask R-CNN:}
Mask R-CNN is a deep learning model widely used for object detection and instance segmentation, known for its accuracy in delineating individual objects by predicting both bounding boxes and segmentation masks \citep{He2020MaskR-CNN}. Building on Faster R-CNN, it features a backbone network for feature extraction, a region proposal network (RPN), and dual branches for object classification and mask prediction. While highly effective, its use in agriculture faces challenges due to environmental variability, dataset dependency, and high computational demands, limiting real-time deployment. Recent research focuses on optimizing Mask R-CNN through adaptive architectures, enhanced data augmentation, and lightweight models. Successful applications include robotic tea-leaf harvesting, strawberry ripeness detection, apple flower identification, and fruit stress monitoring—demonstrating its potential to improve precision agriculture and yield outcomes.

\textbf{Segment anything model (SAM):}
Segment anything model (SAM) introduces a promptable foundation model  and the largest segmentation dataset to date (SA-1B), enabling impressive zero-shot segmentation performance across diverse tasks and image distributions, often rivaling fully supervised methods \citep{Kirillov2023SegmentAnything}. SAM 2 is a promptable foundation model with a streaming memory transformer architecture, designed for efficient and accurate real-time image and video segmentation using a data engine that iteratively improves performance through user interaction \citep{Ravi2024SAMVideos}.

\begin{figure}[h]
 \centering
\includegraphics[width=\linewidth]{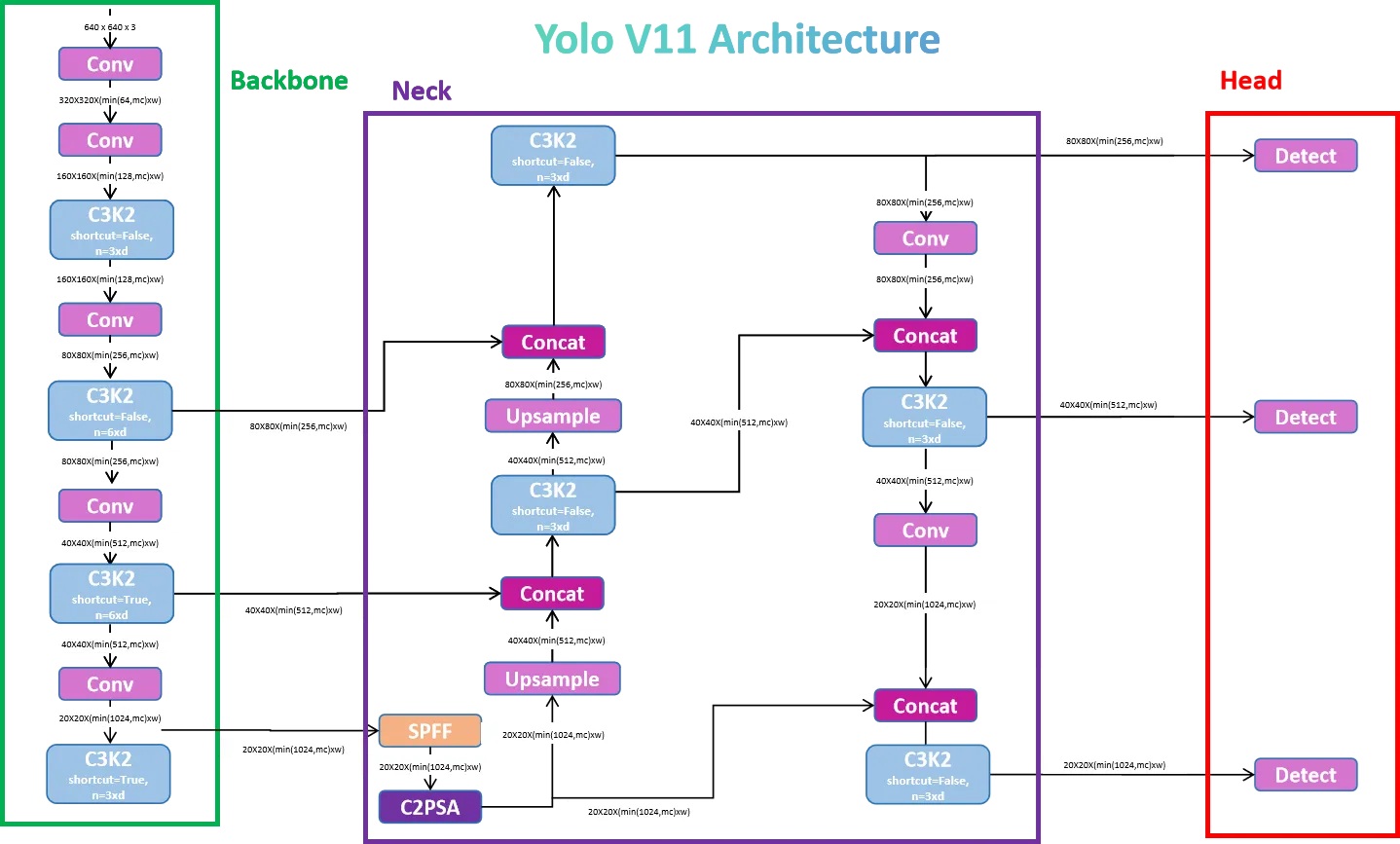}
 \caption{The YOLOv11 network structure.} \url{https://medium.com/@nikhil-rao-20/yolov11-explained-next-level-object-detection-with-enhanced-speed-and-accuracy-2dbe2d376f71} \label{fig:yolo}
\end{figure}

\textbf{YOLO:}
A unified, fast, and real-time object detection framework, YOLO, reformulates detection as a single regression task—mapping directly from image pixels to bounding box coordinates and class probabilities—thereby enabling end-to-end optimization \citep{Redmon2015YouDetection}. YOLOv4 integrates various architectural and training features, including Weighted Residual Connections, Cross-Stage Partial connections, Cross mini-Batch Normalization, Self-Adversarial Training, Mish activation, Mosaic data augmentation, DropBlock regularization, and CIoU loss, achieving real-time inference on the MS COCO dataset \citep{Bochkovskiy2020YOLOv4:Detection}. YOLOv7 delivers state-of-the-art real-time object detection performance, surpassing existing detectors in both speed (5–160 FPS) and accuracy (56.8\% AP at $\geq$30 FPS on V100), without relying on external data or pre-training \citep{Wang2023YOLOv7:Detectors}. The architecture of the YOLOv8 network is primarily composed of three main components: the backbone, neck, and head. The backbone, based on a modified CSPDarknet53, uses the lightweight C2f module for efficient feature extraction and integrates SPPF for adaptive pooling with reduced latency \citep{Wang2023UAV-YOLOv8:Scenarios}. The neck adopts a PAN-FPN structure to fuse deep semantic and shallow positional features, enhancing feature diversity while maintaining model efficiency. The head uses a decoupled structure for classification and bounding box regression, leveraging distinct loss functions (BCE, DFL, CIoU) and an anchor-free design with Task-Aligned Assigner for improved detection accuracy and convergence.  YOLOv9 introduces Programmable Gradient Information (PGI) to enhance gradient flow by preserving input information through network layers and proposes GELAN, a lightweight architecture that leverages PGI to achieve strong object detection results on MS COCO, outperforming even pre-trained models in some cases \citep{Wang2025YOLOv9:Information}. YOLOv11 introduces key architectural enhancements such as the C3k2 block, SPPF, and C2PSA module as illustrated in Figure~\ref{fig:yolo}. These components enhance feature extraction and significantly improve performance across tasks including object detection, instance segmentation, and pose estimation \citep{Khanam2024YOLOv11:Enhancements}.

\section{Case study}       \label{sec:case}
We evaluated the performance of our proposed approach using real-world subsea pipeline inspection datasets. Incorporating domain-specific data, particularly historical inspection patterns and structural condition records, is crucial to enhancing the reliability and applicability of pipeline outlining in underwater.

\subsection{Data description}
\begin{figure}[h]
  \centering
  \includegraphics[width=\linewidth]{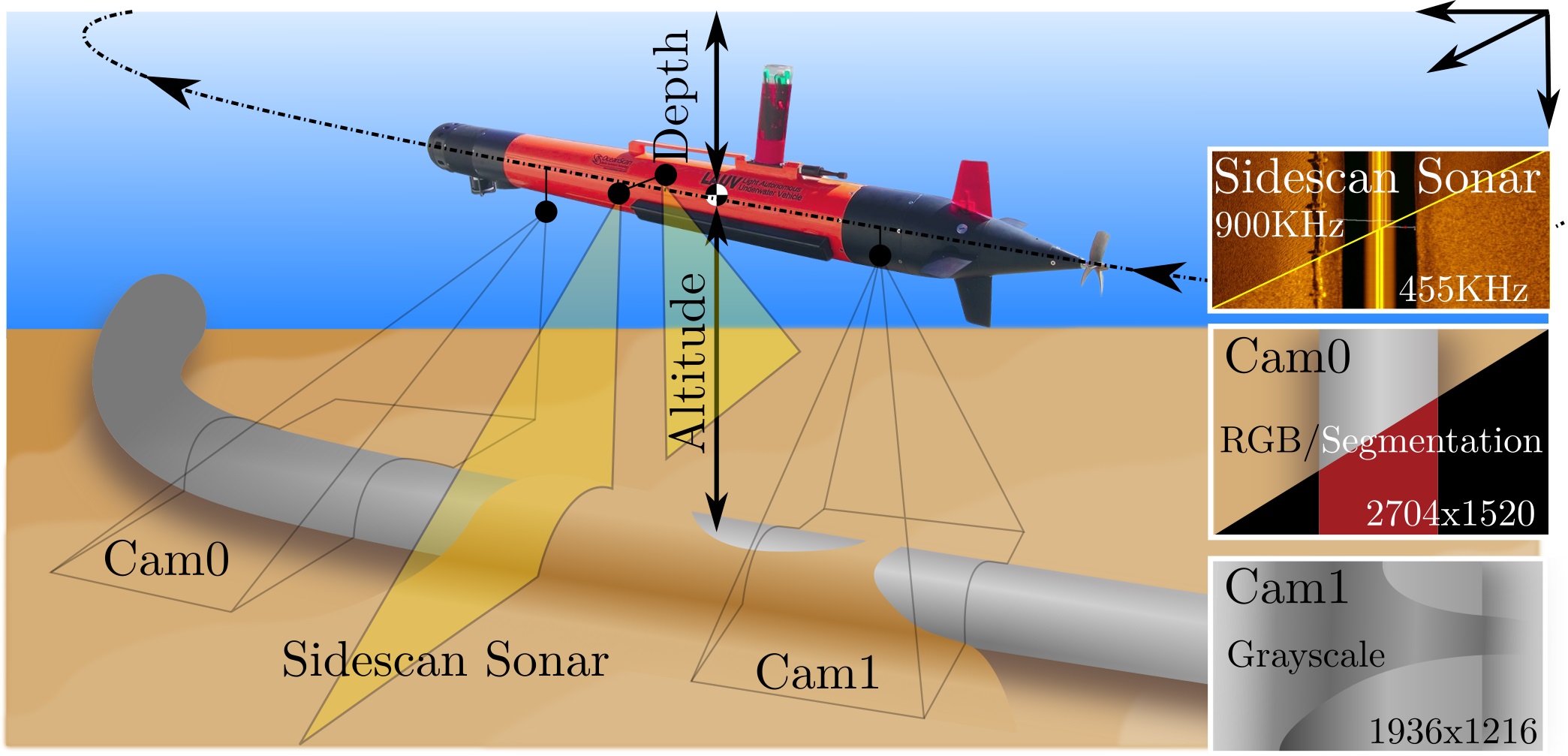}
  \caption{Overview of Submarine pipeline inspection data collection setup. Adapted from \citep{Alvarez-Tunon2024SubPipe:Localization}}  \label{fig:inspection}
\end{figure}

SubPipe is a dataset collected during a survey mission conducted by OceanScan-MST1 along a 1 km underwater pipeline in Porto, Portugal, comprising onboard sensor measurements and the robot's estimated state. Comprehensive details about the experimental data collection platform and dataset description are provided in \citep{Alvarez-Tunon2024SubPipe:Localization}.  Figure~\ref{fig:inspection} shows the experimental setup of the SubPipe dataset collection during an underwater pipeline inspection mission using OceanScan’s LAUV. The data collection setup includes two downward-looking RGB cameras, a Klein 3500 sss scanner, and synchronised sensor data such as temperature, depth, altitude, pose, velocity, and acceleration measurements. SubPipe data comprises five video sequences, each lasting 7–9 minutes, captured using a high-resolution camera at 240 Hz and a low-resolution camera at 4 Hz. Side-scan sonar (SSS) images are annotated in COCO format for object detection and resized to 640 × 640 pixels to ensure algorithm compatibility. The dataset includes ground truth annotations for object detection, image segmentation, and visual-inertial localization, with RGB images semantically segmented and SSS images specifically labeled for object detection, supporting a wide range of research applications. In this work, we use SubPipeMini, as it exclusively provides ground truth annotations for image segmentation.

\subsection{Data preparation and model settings}
SubPipeMini dataset comprising 647 annotated frames. Figure~\ref{fig:steps}  illustrats Workflow of preprocessing of underwater images, including format conversion and dataset splitting; followed by model training, inference on the test set, and evaluation to identify the best-performing model. The data preparation begins with preprocessing underwater image frames by converting them all to JPG format and renaming them numerically. The sorted images and labels are renamed using numbering and organized them into separate folders for training, validation, and testing. The first 60\% of labeled images were allocated for training, 20\% for validation, and the remaining 20\% for testing. The dataset is then divided into training, validation, and test sets. Multiple segmentation models are trained on the assigned datasets, and the best-performing model is used to perform inference on the test set. Finally, evaluation metrics are computed for each model to compare performance and identify the most effective model for underwater image segmentation. 

\usetikzlibrary{arrows.meta, positioning}

\tikzstyle{block} = [
    rectangle, draw, align=center,
    text width=24em,        
    minimum height=2em,     
    inner sep=2pt           
]
\tikzstyle{arrow} = [thick, ->, >=stealth]

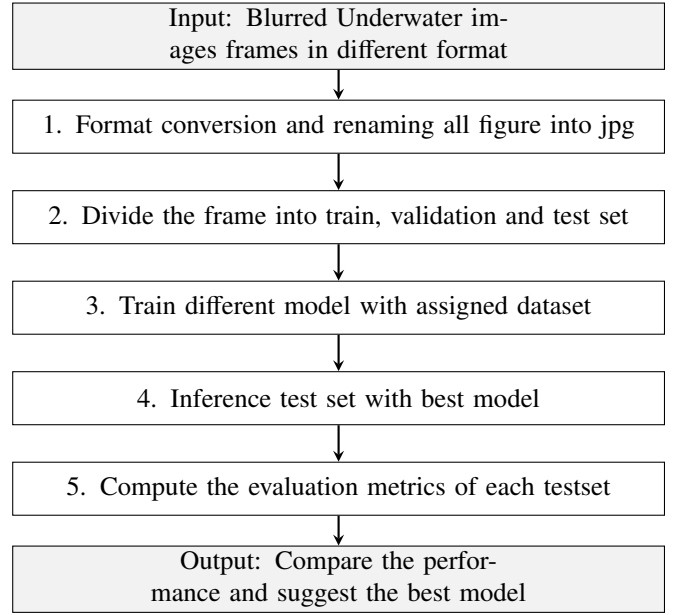
\begin{figure}[htbp]
\begin{center}
\begin{tikzpicture}[node distance=1.2cm]

\node (input) [block, fill=gray!10] {Input: Blurred Underwater images frames in different format };
\node (enhance) [block, below of=input] {1. Format conversion  and renaming all figure into jpg };
\node (detect) [block, below of=enhance] {2. Divide the frame into train, validation and test set};
\node (slam) [block, below of=detect] {3. Train different model with assigned dataset};
\node (reconstruct) [block, below of=slam] {4. Inference test set with best model};
\node (defect) [block, below of=reconstruct] {5. Compute the evaluation metrics of each testset};
\node (output) [block, below of=defect, fill=gray!10] {Output: Compare the performance and suggest the best model};

\draw [arrow] (input) -- (enhance);
\draw [arrow] (enhance) -- (detect);
\draw [arrow] (detect) -- (slam);
\draw [arrow] (slam) -- (reconstruct);
\draw [arrow] (reconstruct) -- (defect);
\draw [arrow] (defect) -- (output);

\end{tikzpicture}
\end{center}
\caption{Block diagram of the YOLO based image segmentation workflow illustrating preprocessing, dataset preparation, model training, inference, and performance evaluation.}
\label{fig:steps}
\end{figure}

\subsection{Computation setup}
All segmentation models were trained on a workstation with an Intel(R) Core(TM) i9-10900X CPU @ 3.70GHz CPU, NVIDIA T600 GPU, 31.7 GB RAM, and Windows 11 Education using PyTorch training environment CUDA 12.1. Training used a batch size of 8, an initial learning rate of 0.001 with momentum 0.937, weight decay 0.0005, and a dropout rate of 0.5 to prevent overfitting. Early stopping was applied if validation performance did not improve for 5 consecutive epochs during up to 50 iterations, with a 3-epoch warm-up phase to stabilize optimization. The model is trained using the Adam optimizer with input images resized to 640 × 640 pixels.

\subsection{Evaluation metrics}       \label{sec:metrics}
The performance is assessed using four standard segmentation metrics as in \citep{Dang2024Two-layerSegmentation}: Dice coefficient, Intersection-over-Union (IoU), Hausdorff Distance (HD), and Mean Absolute Distance (MAD) as in Equations~\ref{eq:dice}–~\ref{eq:mad}. The mean Intersection over Union (mIoU) is the average IoU across all classes and provides an overall performance measure for multi-class segmentation tasks. Since our work segmentation involves a single foreground class (pipe) against the background, the mIoU is equivalent to the IoU. Low values of HD and MAD, or high values of Dice coefficient and mIoU, indicate better segmentation quality. Let there be $C$ classes and $N$ images, each of size $H \times W$. Let $\hat{y}_{c} \in \{0,1\}^{N \times H \times W}$ and $y_{c} \in \{0,1\}^{N \times H \times W}$ denote the predicted and ground truth binary masks for class $c$, respectively. Then:

\begin{itemize}
    \item The Dice coefficient for class $c$ is defined as:
    \begin{equation}
        \text{Dice}_c = \frac{2 \sum \hat{y}_c y_c}{\sum \hat{y}_c + \sum y_c}  \label{eq:dice}
    \end{equation}

    \item The Intersection-over-Union (IoU) or Jaccard Index for class $c$ is:
    \begin{equation}
        \text{IoU}_c = \frac{\sum \hat{y}_c y_c}{\sum \hat{y}_c + \sum y_c - \sum \hat{y}_c y_c}
    \end{equation}
    where the final IoU score is the average over all $C$ classes.

    \item The Hausdorff Distance (HD) measures the maximum distance between the ground truth contour $G_c$ and the predicted contour $P_c$ for class $c$:
    \begin{equation}
        \text{HD}_c = \max \left\{ \sup_{g \in G_c} \inf_{p \in P_c} \|g - p\|, \sup_{p \in P_c} \inf_{g \in G_c} \|p - g\| \right\}
    \end{equation}

    \item The Mean Absolute Distance (MAD) for class $c$ is:
    \begin{equation}
        \text{MAD}_c = \frac{1}{|G_c| + |P_c|} \left( \sum_{g \in G_c} \inf_{p \in P_c} \|g - p\| + \sum_{p \in P_c} \inf_{g \in G_c} \|p - g\| \right)   \label{eq:mad}
    \end{equation}
\end{itemize}

\section{Results and discussion}       \label{sec:result}

\subsection{Image enhancement}       \label{sec:image_ehnan}
\begin{figure}[h]
  \centering
  \includegraphics[width=\linewidth]{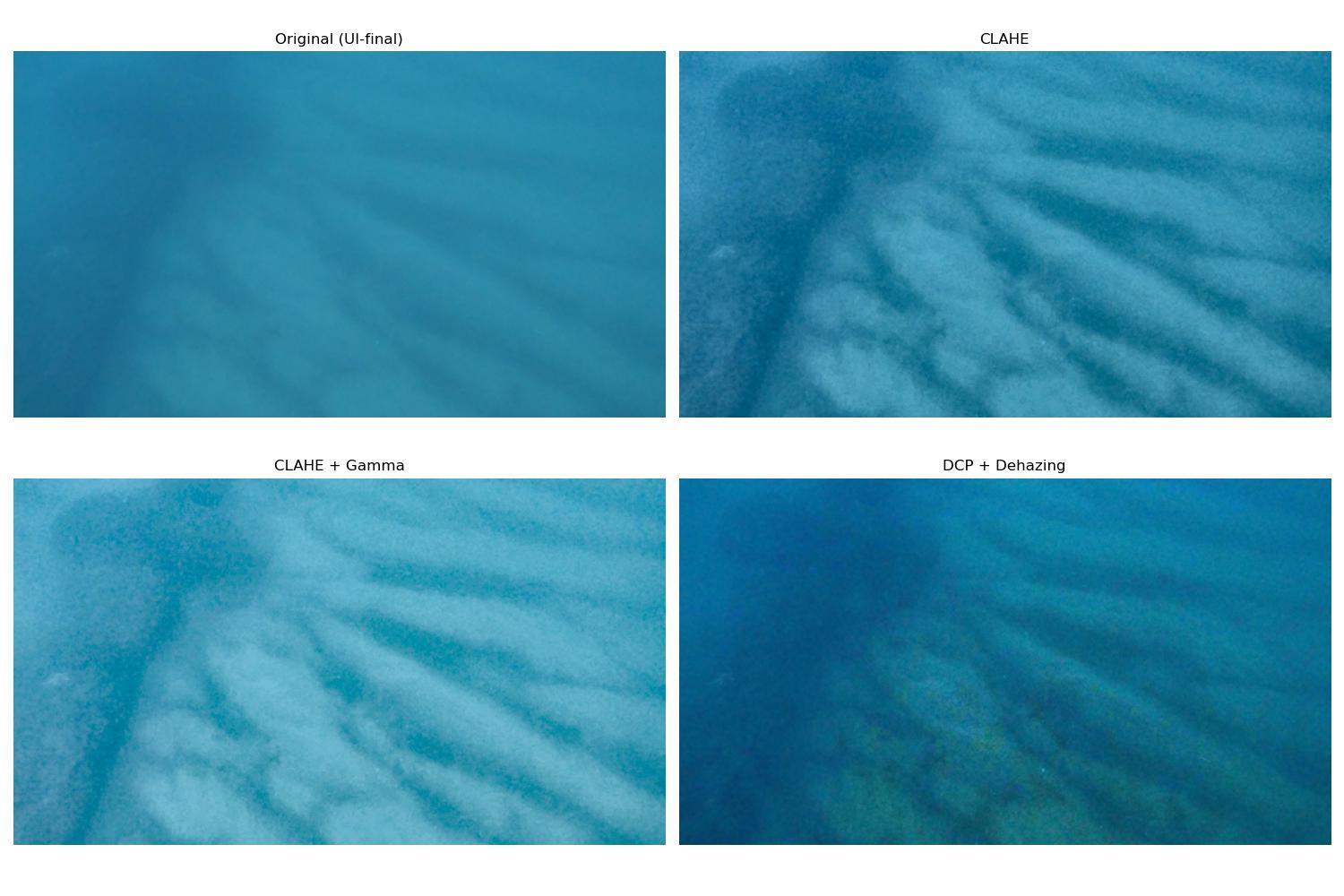}
  \caption{Comparative Visualization of Image Enhancement Techniques.}  \label{fig:enhanced}
\end{figure}

Two image pre-processing techniques  CLAHE and DCPD are tested aiming to enhance contrast or visibility but can distort image features critical for accurate segmentation. Figure~\ref{fig:enhanced} presents a 2×2 comparison of image enhancement methods applied to an underwater pipeline image. The top-left image shows the original input (UI-final). The top-right applies CLAHE to improve local contrast. The bottom-left further enhances this using gamma correction for brightness balancing. The bottom-right displays the result of a Dark Channel Prior (DCP) combined with a dehazing algorithm, significantly improving visibility and contrast by removing haze and color distortions. Together, these subplots illustrate progressive enhancement in image clarity and detail. These enhancements may introduce artifacts or alter the pixel-level context, leading to reduced performance in boundary-sensitive metrics such as HD and MAD. As a result, segmentation models may struggle to adapt to the modified data, producing less accurate predictions. Perhaps a more effective alternative could be domain-specific data augmentation could preserve the natural image structure while improving feature learning.

\subsection{Training performance}       \label{sec:train}

\begin{figure}[h]
  \centering
  \includegraphics[width=\linewidth]{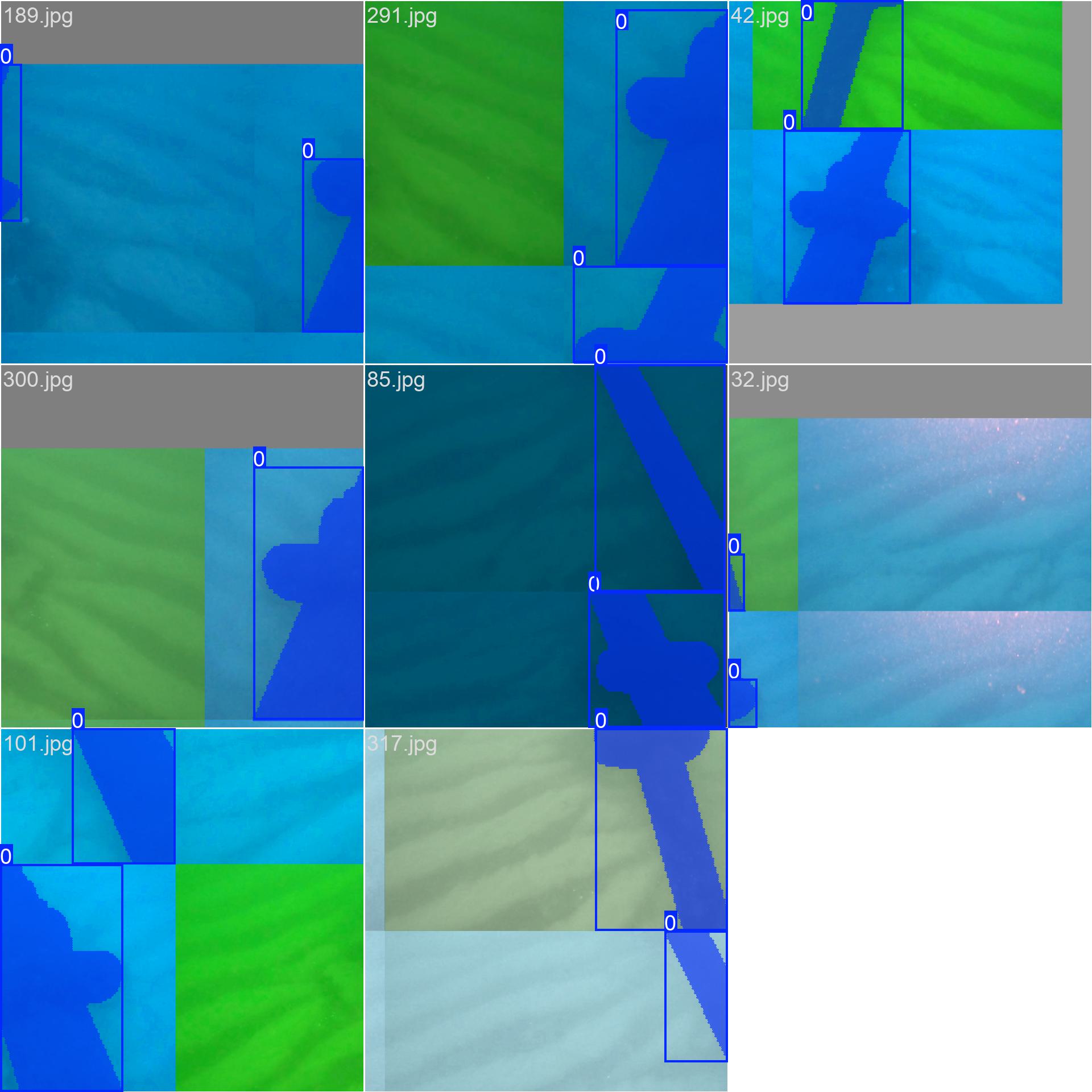}
  \caption{Training batch visualization showing bounding boxes and segmentation masks for subsea pipeline detection using YOLOv11n.}  \label{fig:train_batch}
\end{figure}

Segmented batch image results provide pixel-wise predictions for each image in the batch, allowing simultaneous visualization and evaluation of multiple outputs. Figure~\ref{fig:train_batch} illustrates a batch of training samples used to train the YOLOv11n segmentation model for subsea pipeline inspection. Each tile represents an image containing subsea scenes where pipelines are partially or fully visible. The overlaid blue regions show the predicted segmentation masks, while the blue bounding boxes highlight the localized pipe objects, both labeled with class ID "0" indicating the pipeline class. The presence of masks and bounding boxes confirms that the model is being trained simultaneously for object detection (bounding box localization) and pixel-wise segmentation (mask generation). The diversity in lighting, texture, and background conditions reflects the challenges inherent in underwater imaging and demonstrates the model’s ability to learn robust features across varied subsea environments.

\begin{figure}[h]
  \centering
  \includegraphics[width=\linewidth]{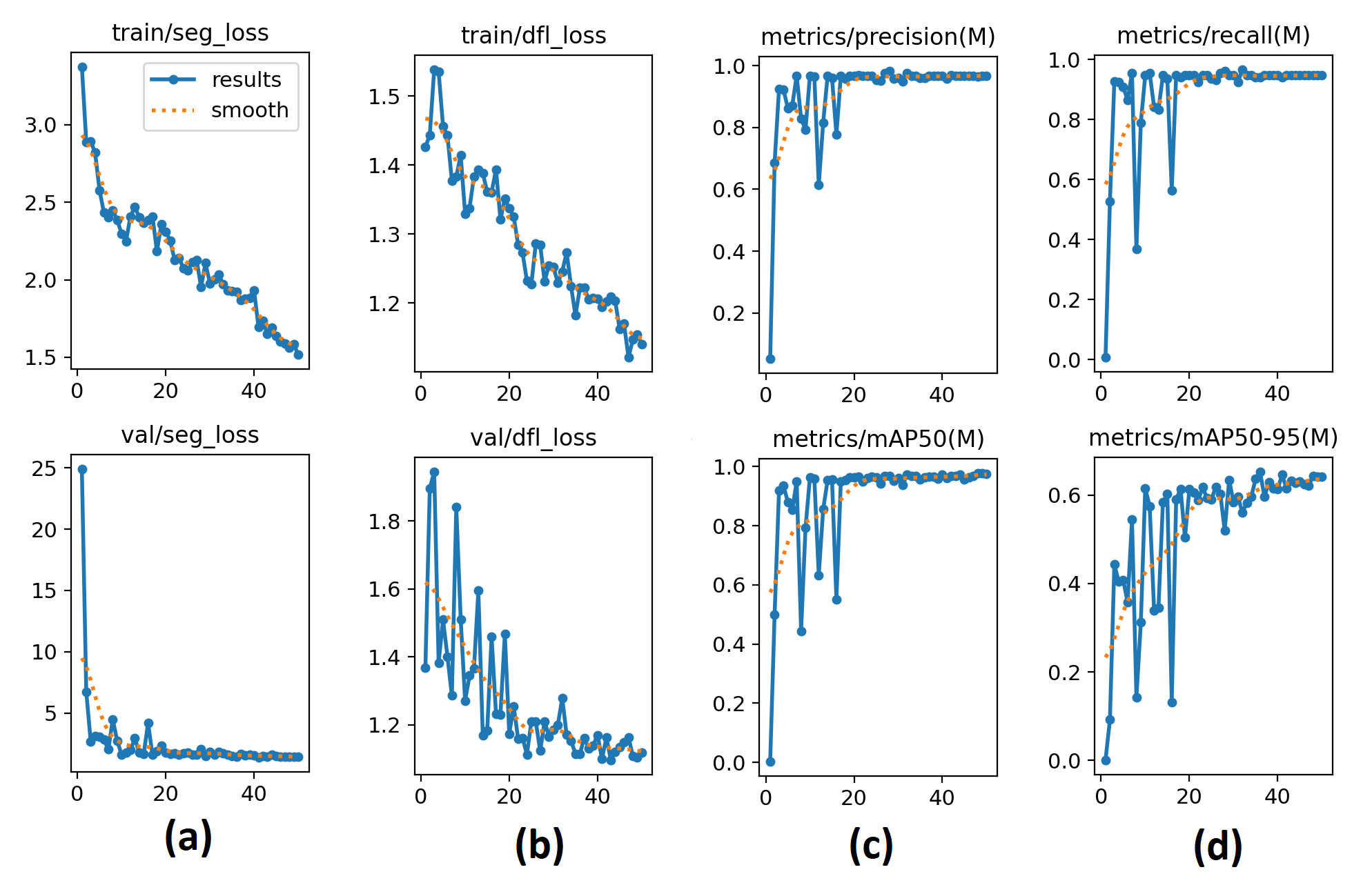}
  \caption{Training and validation performance curves of the YOLOv11n segmentation model on subsea pipeline inspection data.}  \label{fig:result}
\end{figure}

Figure~\ref{fig:result} presents the training and validation performance curves of the YOLOv11n segmentation model applied to subsea pipeline inspection imagery. The network is trained for 50 epochs using evolved hyperparameters, with validation at each epoch to assess performance and prevent overfitting. Figure~\ref{fig:result} (a) (top row) illustrates a decreasing trend in the training segmentation losses, indicating effective model convergence, while the corresponding validation losses (bottom row) also show a consistent decline, reflecting good generalization performance. Similarly, Figure~\ref{fig:result} (b) (top row) presents a downward trend in the training Distribution Focal Loss (DFL), with the associated validation losses (bottom row) mirroring this pattern, further supporting the model's ability to generalize well to unseen data. The precision and recall metrics for segmentation masks (M) rapidly approach 1.0, reflecting high accuracy in pixel-wise segmentation in Figure~\ref{fig:result} (c) and (d), respectively. Furthermore, the mean Average Precision (mAP) metrics—mAP@0.5:0.95—for mask predictions show progressive improvement, with final values nearing optimal levels. This suggests that the YOLOv11n model demonstrates strong capability in accurately detecting and segmenting pipelines under challenging underwater visual conditions.

\subsection{Validation performance}       \label{sec:valid}

\begin{figure*}[h]
  \centering
  \includegraphics[width=\linewidth]{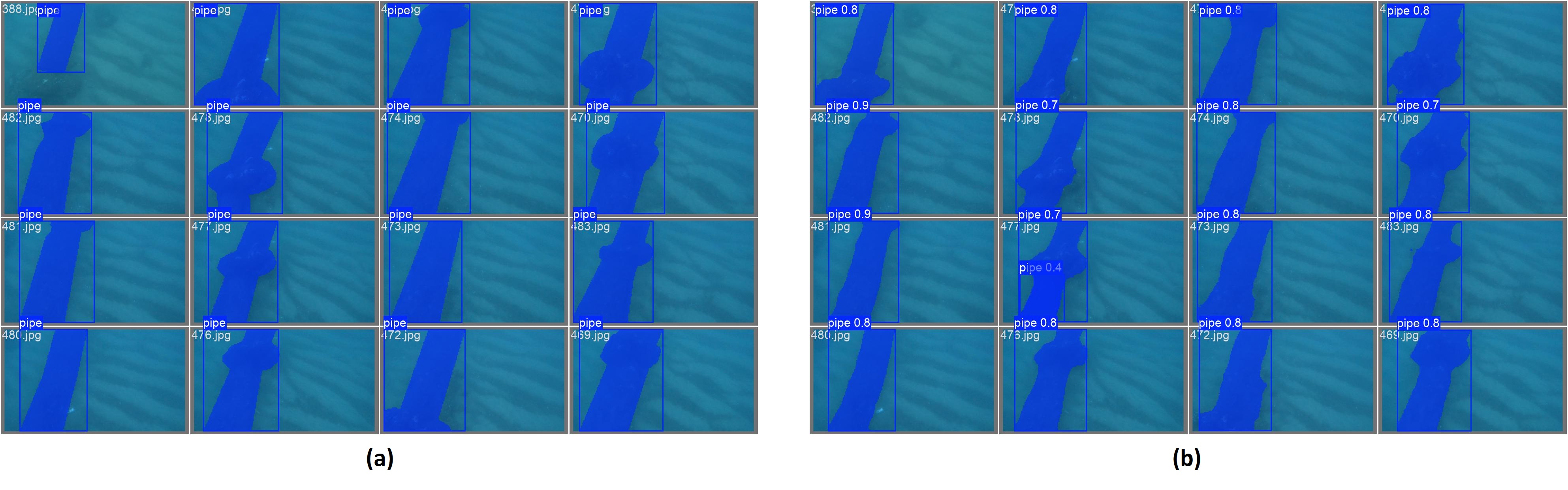}
  \caption{Validation batch visualization showing bounding boxes and segmentation masks for subsea pipeline detection using YOLOv11n.}  \label{fig:val_batch}
\end{figure*}

Figure~\ref{fig:val_batch} illustrates the validation batch outputs for a YOLOv11n segmentation model trained on subsea pipeline imagery. The plot on the left of Figure~\ref{fig:val_batch} (a) shows the ground truth labels with precise masks and bounding boxes for the "pipe" class, which serves as the reference standard. The right plot in Figure~\ref{fig:val_batch} (b) shows the corresponding model predictions, where the predicted pipeline masks and bounding boxes are overlaid along with confidence scores. Visual comparison reveals that the model accurately detects pipes in all images, while segmenting them in most images with slight variations in mask shapes and confidence levels, indicating good alignment between the ground truth and predictions in the validation phase.

Training requires storing activations, gradients, and optimizer states, which significantly increases GPU memory consumption. To prevent out-of-memory errors, the training batch size is often reduced (e.g., to 9). In contrast, validation only involves forward passes without backpropagation, resulting in lower memory usage and enabling a larger batch size (e.g., 16) for faster processing.

After training the model, two distinct performance curves are obtained: the Box F1 Curve, which evaluates the model’s capability in object detection by measuring the accuracy of predicted bounding boxes, and the Mask F1 Curve, which assesses the model’s effectiveness in pixel-wise segmentation by quantifying how well the predicted masks align with the ground truth. These curves provide complementary insights—while the box curve reflects spatial localization performance, the mask curve captures the precision of boundary delineation essential for fine-grained segmentation tasks.

\begin{figure}[h]
  \centering
  \includegraphics[width=\linewidth]{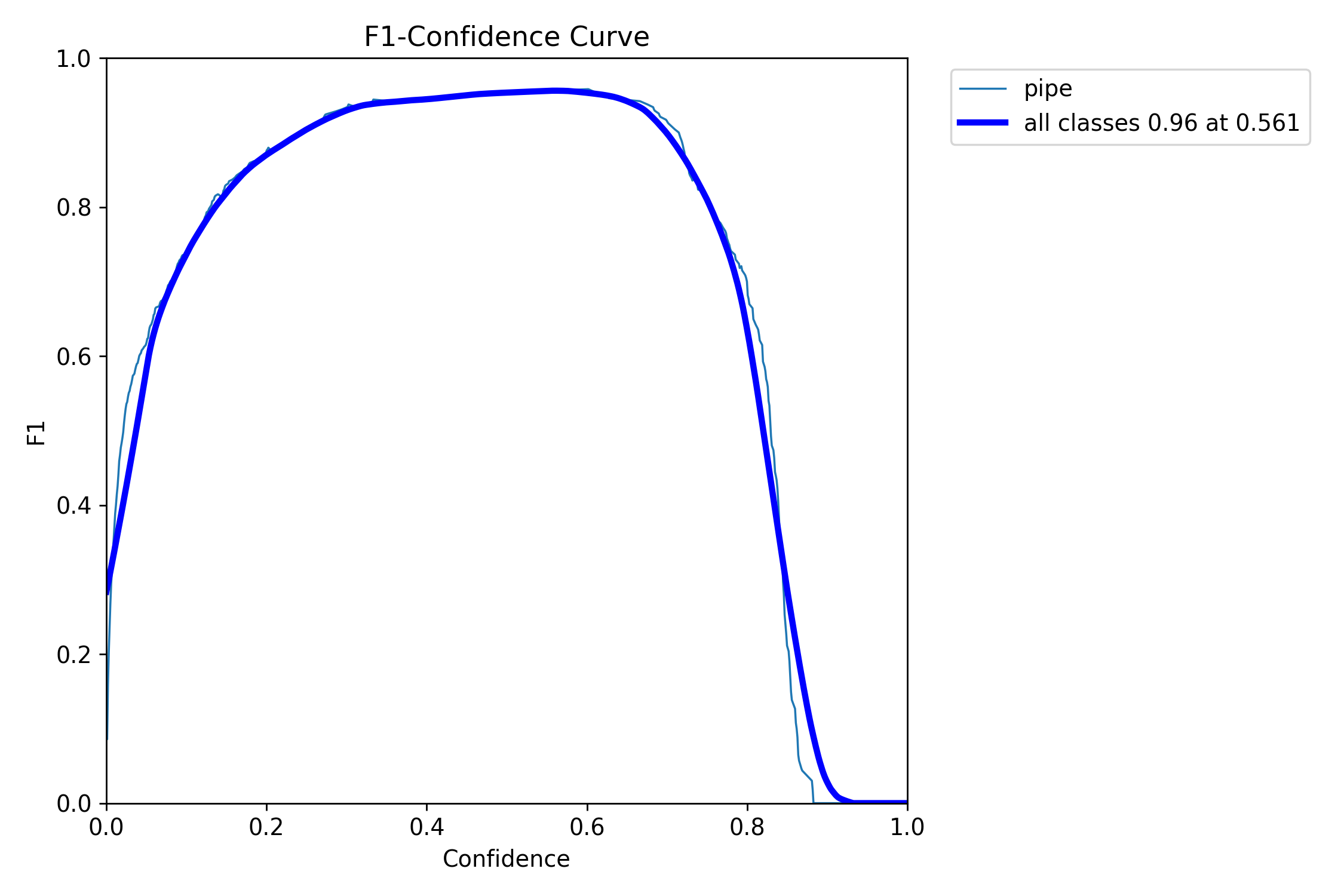}
  \caption{F1-confidence curves illustrating the trade-offs and reliability of segmentation predictions across varying thresholds.}  \label{fig:curve1}
\end{figure}

Figure~\ref{fig:curve1} shows how the F1-score (harmonic mean of precision and recall) changes as you adjust the confidence threshold used to accept predictions. An F1-score of 0.96 with an optimal confidence threshold of 0.561 indicates the model achieves a strong precision-recall balance, maximizing accurate and complete predictions. YOLO segmentation model shows strong and balanced performance across all classes.  Meanwhile, the mask F1-confidence curve peaks slightly higher, at an F1-score of 0.80 at a threshold of 0.516, suggesting that the model is slightly better at producing accurate segmentation masks than just detecting objects. Since both curves peak at nearly the same threshold and all classes share the same F1 performance, this reflects a well-trained model that generalizes consistently across the dataset. You can confidently use a threshold around 0.515–0.516 for both detection and segmentation tasks.

\begin{figure}[h]
  \centering
  \includegraphics[width=\linewidth]{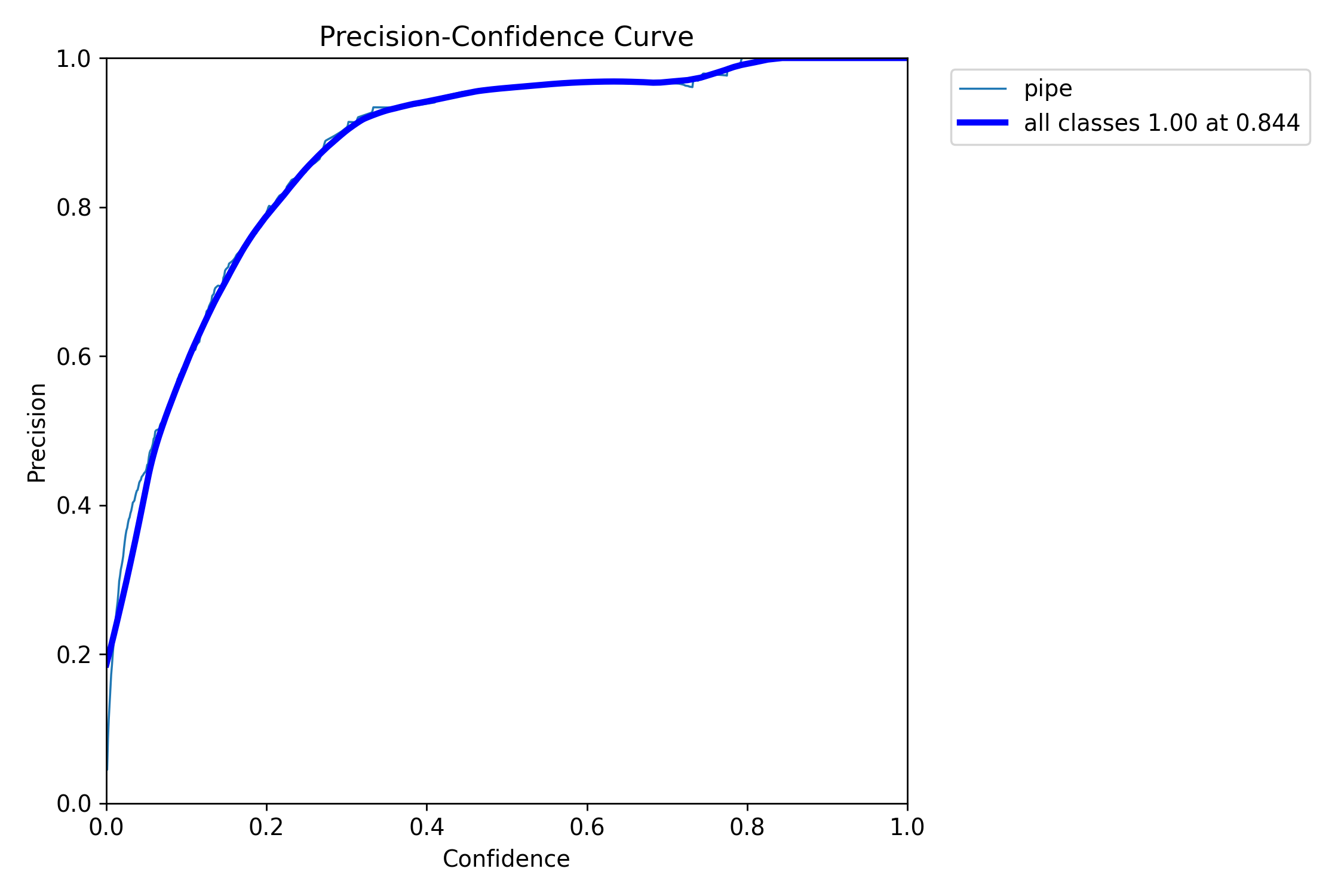}
  \caption{Precision-confidence curves illustrating the trade-offs and reliability of segmentation predictions across varying thresholds.}  \label{fig:curve2}
\end{figure}

Figure~\ref{fig:curve2} shows how model precision varies with different confidence score thresholds assigned to predicted segments. It helps assess the reliability of predictions—higher confidence scores should ideally correspond to higher precision, indicating well-calibrated model outputs. The mask precision-confidence curve shows a precision score of 1.0 for all classes at a confidence threshold of 0.8444. This indicates that at this threshold, every predicted mask is correct—there are no false positives, which suggests the model is highly precise at this confidence level, but likely making very few predictions (low recall). It is extremely cautious, only predicting when it is very sure, which is excellent for applications where false positives are costly, such as medical image segmentation and pipeline safety measures.

\begin{figure}[h]
  \centering
  \includegraphics[width=\linewidth]{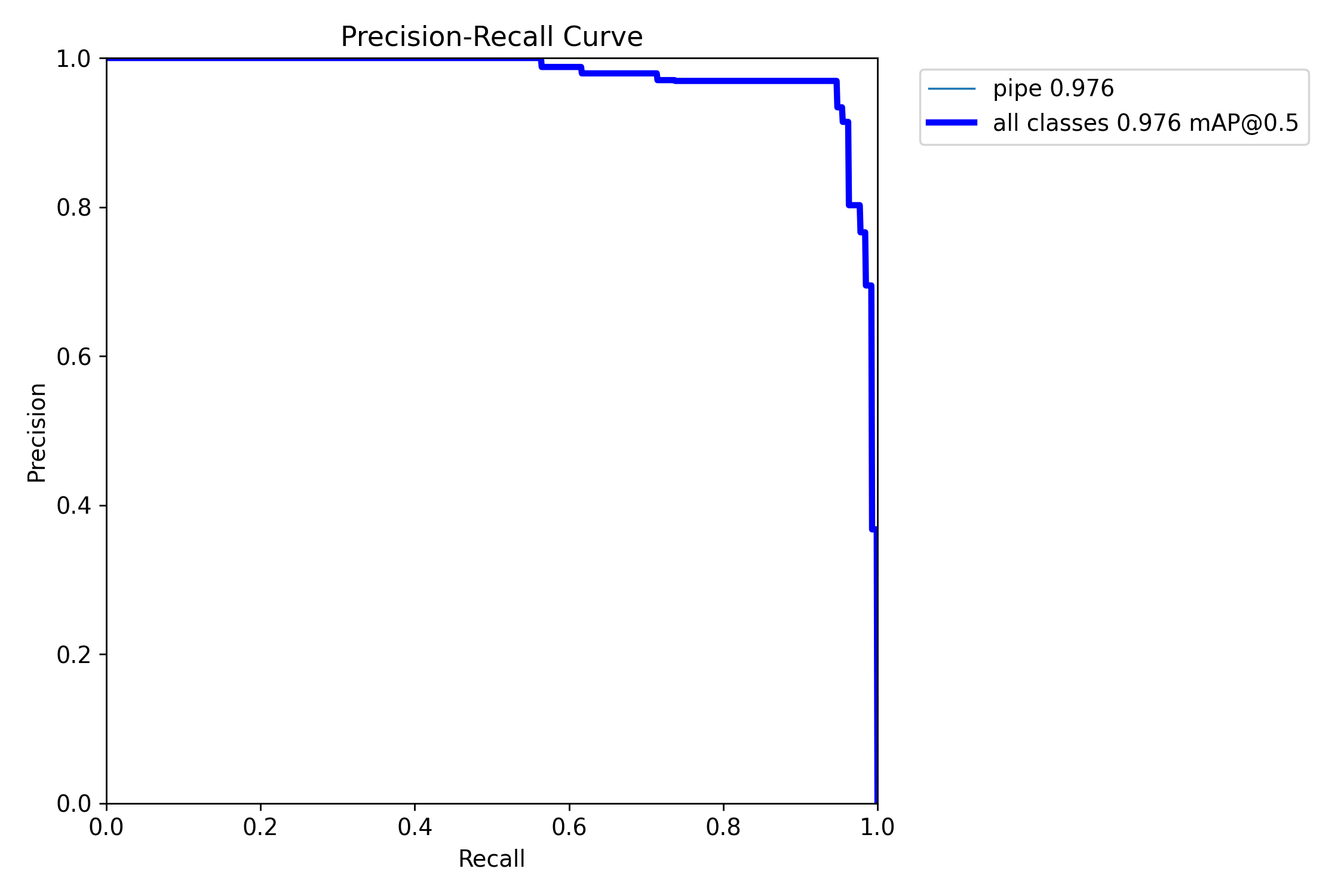}
  \caption{Precision-recall curves illustrating the trade-offs and reliability of segmentation predictions across varying thresholds.}  \label{fig:curve3}
\end{figure}

Figure~\ref{fig:curve3} illustrates the trade-off between precision (the accuracy of predicted segment pixels) and recall (the completeness of those predictions) across different threshold values. It is beneficial in evaluating segmentation performance on imbalanced datasets, where correctly identifying minority classes (e.g., small or thin pipeline regions) is critical. The mask precision-recall curve shows a mean Average Precision (mAP@0.5) of 0.976 for the "pipe" class and also for all classes, indicating consistent and accurate segmentation performance.  The model performs very well and consistently across all classes, with slightly better segmentation accuracy than bounding box detection. The close alignment of class-specific and overall scores suggests a well-balanced dataset and good generalisation. The results are reliable for both detection and segmentation tasks, with powerful performance on the pipe class.

\begin{figure}[h]
  \centering
  \includegraphics[width=\linewidth]{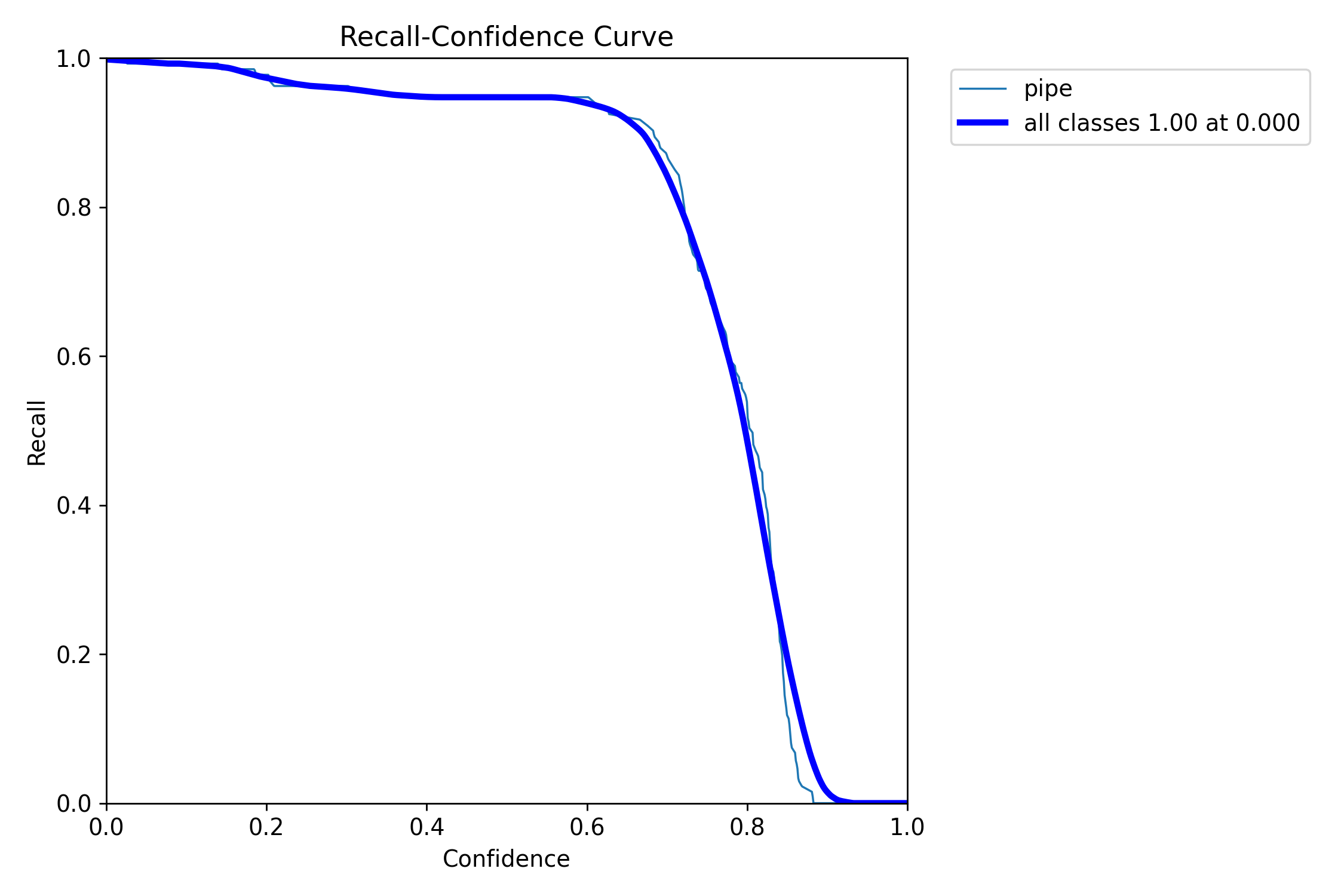}
  \caption{Recall-confidence curves illustrating the trade-offs and reliability of segmentation predictions across varying thresholds.}  \label{fig:curve4}
\end{figure}

Figure~\ref{fig:curve4} depicts how recall changes as the confidence threshold for predictions increases. It helps evaluate the model's ability to detect all relevant segments at varying confidence levels, with lower thresholds typically capturing more true positives but also potentially more false positives. The mask recall-confidence curve shows a recall of 1 for all classes at a confidence threshold of 0.0. This means that when the model accepts all predictions regardless of confidence, it correctly detects 100\% of all true objects for segmentation masks. At zero confidence threshold, the model maximises recall by including even low-confidence predictions, capturing almost all true objects. However, this likely comes with many false positives. This curve illustrates the model’s strong ability to identify true positives, but highlights the need to increase the confidence threshold to balance recall with precision for practical use.

\subsection{Testing performance}       \label{sec:test}

Table~\ref{tab:perfor} shows the segmentation performance of various YOLOv8-seg and YOLOv11-seg model variants. Among the YOLOv8 variants, YOLOv8s achieved the highest mIoU (0.6950) and Dice score (0.8009), indicating superior segmentation accuracy compared to YOLOv8n and YOLOv8m. However, the YOLOv11n model outperformed all others with the highest mIoU (0.7098), Dice score (0.8129), and the lowest MAD (24.94), demonstrating a strong balance between accuracy and contour precision. YOLOv11s also showed competitive performance with slightly lower metrics than YOLOv11n but better HD and MAD than YOLOv8 counterparts. In contrast, YOLOv11m yielded the lowest mIoU and Dice scores among the YOLOv11 models and exhibited relatively higher HD and MAD, suggesting reduced performance in this specific segmentation task. Overall, the results highlight the effectiveness of the YOLOv11n variant for underwater pipeline segmentation, offering improved segmentation quality and boundary alignment.

\begin{table}
  \caption{Segmentation performance comparison of various variants of YOLOv8-seg and YOLOv11-seg models on underwater pipeline}    \label{tab:perfor}
  \begin{tabular}{lccccc}
    \toprule
    Model & mIoU & Dice & HD & MAD \\
    \midrule
    YOLOv8n      & 0.6673  & 0.7689  & 383.46  & 53.39  \\   
    YOLOv8s      & 0.6950  &  0.8009  & 322.72   & 30.29  \\ 
    YOLOv8m      & 0.6823  &  0.7902 &  338.52 &   30.99  \\
    YOLOv11n     & 0.7098  & 0.8129  & 307.59  & 24.94  \\
    YOLOv11s     & 0.7042  & 0.8088  & 304.60  & 25.15  \\
    YOLOv11m     & 0.6722  & 0.7780  & 352.10  & 36.49  \\
    \bottomrule
  \end{tabular}
\end{table}

Table~\ref{tab:perfor2} shows the impact of image enhancement techniques—CLAHE  and DCPD on segmentation performance was evaluated using YOLOv8n and YOLOv11n models across metrics: mIoU, Dice coefficient, HD, and MAD. For YOLOv8n, applying CLAHE resulted in a marginal improvement in mIoU (0.6688 vs. 0.6673) and Dice score (0.7717 vs. 0.7689), but also led to increased HD and MAD, indicating more dispersed boundary predictions. Conversely, DCPD led to a decline in mIoU and Dice (0.6368 and 0.7563, respectively), with a higher HD but a moderately improved MAD. For YOLOv11n, the original model achieved the best overall performance (mIoU: 0.7098, Dice: 0.8129, HD: 307.59, MAD: 24.94). Enhancement via CLAHE slightly reduced segmentation accuracy and  worsened MAD (31.52), while DCPD similarly decreased all performance metrics. These results suggest that while CLAHE can offer minor benefits in certain scenarios, both enhancement techniques may degrade segmentation accuracy and spatial precision, particularly when using the more capable YOLOv11n model, which performs best on unprocessed images.


\begin{table}
  \caption{Segmentation performance comparison of original image with preprocessed image by CLAHE and DCPD techniques}    \label{tab:perfor2}
  \begin{tabular}{lcccccc}
    \toprule
    Model & Enhancement & mIoU & Dice & HD & MAD \\
    \midrule
    
     & Original  & 0.6673  & 0.7689  & 383.46  & 53.39  \\
   YOLOv8n  & CLAHE   & 0.6688  & 0.7717  & 399.87  & 61.73  \\
    & DCPD  &  0.6368  &  0.7563  & 406.05  & 42.53  \\
    \midrule
  
     & Original    & 0.7098  & 0.8129  & 307.59  & 24.94  \\
   YOLOv11n &  CLAHE   &  0.7048   &  0.8077   &  320.22  & 31.52   \\
    & DCPD   & 0.6749   & 0.7875  & 352.27   & 30.60  \\
    
    \bottomrule
  \end{tabular}
\end{table}

\begin{figure}[h]
  \centering
  \includegraphics[width=\linewidth]{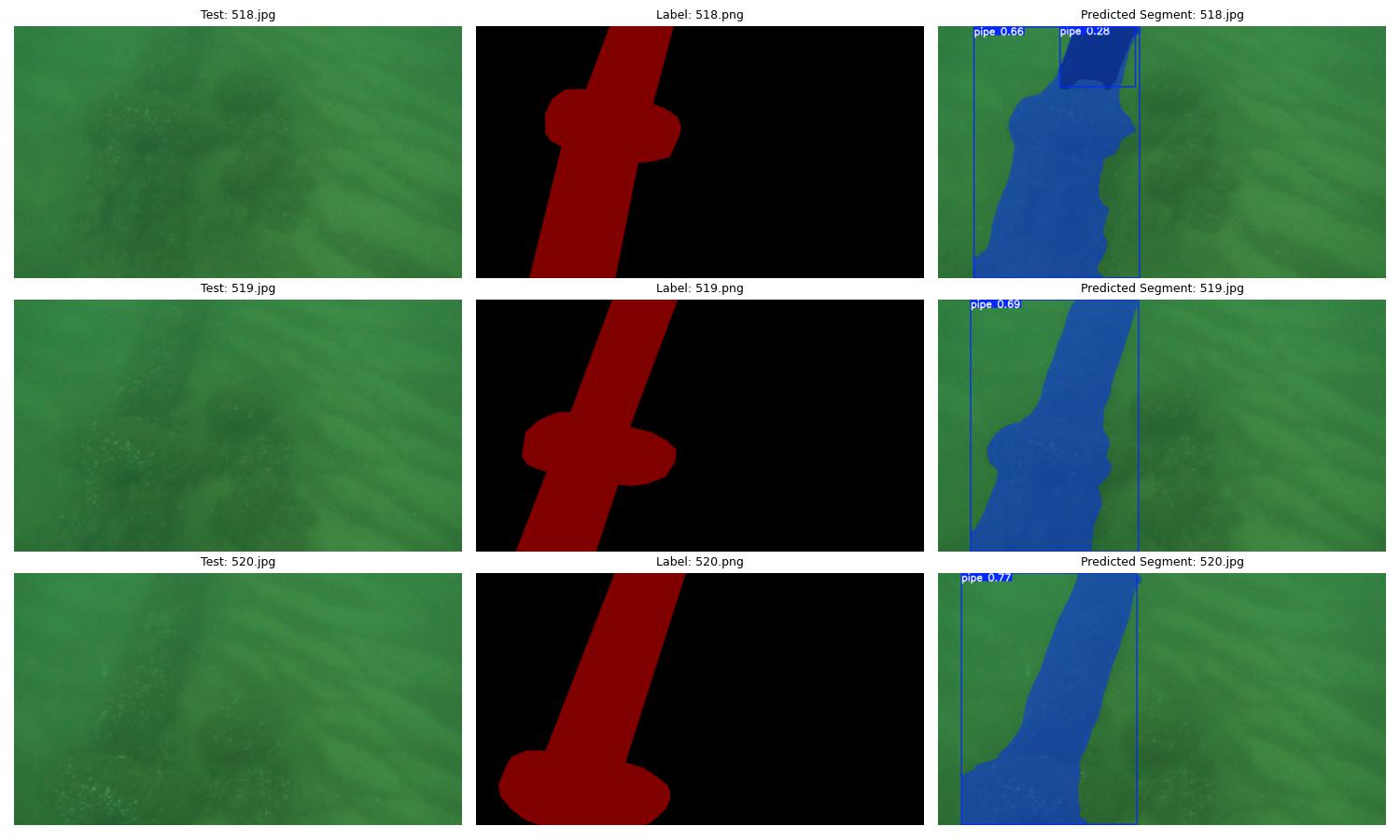}
  \caption{Visual comparison of input images, annotated labels, and YOLOv11n segmentation outputs demonstrating performance on subsea pipeline inspection dataset.}  \label{fig:test_batch}
\end{figure}

Figure~\ref{fig:test_batch} presents a visual comparison of the first three subsea pipeline test images, corresponding ground truth segmentation masks, and the predictions generated by the YOLOv11n model. The first column displays raw test images captured under challenging underwater conditions, such as low contrast, turbidity, and background clutter. The second column displays manually annotated labels that highlight the precise pipeline contours used as ground truth for model evaluation. The third column illustrates the model's predicted segmentations, which closely approximate the annotated masks, demonstrating YOLOv11n’s ability to identify and delineate pipeline regions despite visual noise effectively. The consistency between the predicted and true masks suggests strong model generalization and pixel-wise accuracy, which is essential for reliable subsea infrastructure monitoring and fault detection. This visual evidence supports the quantitative performance metrics and confirms the suitability of YOLOv11n for real-time underwater segmentation tasks.

\subsection{Discussion}       \label{sec:discus}

\textbf{Optimum batch size:}
Varying the batch size can significantly influence segmentation model performance by affecting the stability of gradient updates and the efficiency of memory utilization during training. Smaller batch sizes may promote better generalization in complex datasets, while larger batches benefit from more stable learning but require careful tuning to avoid overfitting or underutilization of model capacity.

\textbf{Image enhancement:}
The evaluation indicates that image enhancement techniques like CLAHE and DCPD provide limited or negative impact on segmentation accuracy and spatial precision for both YOLOv8n and YOLOv11n models. Notably, the YOLOv11n model achieves its best performance on original, unprocessed images, suggesting that preprocessing may not always be beneficial for advanced segmentation networks.

\textbf{Image augmentation:}
Despite the application of standard YOLO augmentations, there remains a need for more task-specific image augmentation strategies tailored to the unique challenges of underwater pipeline imagery. Incorporating geometric distortions, synthetic occlusions, or domain-adaptive noise models could better simulate real-world conditions and improve model generalization. Additionally, advanced augmentation techniques such as generative adversarial networks (GANs) or style transfer may further enrich training data diversity and robustness.

\textbf{Large variants may underperform due to overfitting:}
The larger YOLOv11m variant may exhibit lower segmentation performance than the lighter YOLOv11n due to overfitting on limited or domain-specific underwater data, reducing its generalization to fine-grained segmentation details.

\section{Conclusion and future works}       \label{sec:conclu}
In recent years, there has been growing interest in applying advanced computer vision techniques to subsea infrastructure inspection, particularly for tasks such as underwater pipeline monitoring and maintenance. This study provides a comparative evaluation of tow version of three variants   —YOLOv8, and YOLOv11 models instance segmentation of blurred underwater subsea pipeline images from the SubPipe dataset. Based on the experimental results, the following conclusions were drawn:

All models demonstrated the ability to outline pipelines and surrounding structures despite the inherent challenges of underwater imaging (e.g., blurriness, low contrast, and visual noise). YOLOv11n performed better showing robustness to challenging environmental conditions.

In detecting continuous pipeline structures and small anomalies such as joints or occlusions, YOLOv11n provided higher recall rates, but YOLOv8n produced cleaner segmentation masks with fewer false positives. YOLOv8 exhibited the fastest inference, making it suitable for real-time pipeline inspection systems.
Since CLAHE and DCPD offer limited or negative improvements in segmentation performance, there is a clear need for more advanced image preprocessing techniques as future work to enhance segmentation accuracy in challenging visual environments.

\section*{Acknowledgement} The authors would like to thank Andrzej Wasowski and Olaya Alvarez Tunon from the IT University of Copenhagen, Ranjan Sapkota from Cornell University, and Alexander Lundervold from MMIV, Haukeland University Hospital, for their valuable discussions and insightful suggestions on this work.


\bibliographystyle{elsarticle-harv} 
\bibliography{references}






\end{document}